\pdfoutput=1

\documentclass[11pt]{article}
\usepackage[final]{acl}
\usepackage{xcolor, colortbl} 
\usepackage{array} 

\usepackage{graphicx}
\usepackage{microtype}
\usepackage{hyperref}
\usepackage{url}
\usepackage{booktabs}
\definecolor{darkblue}{rgb}{0, 0, 0.5}
\hypersetup{colorlinks=true, citecolor=darkblue, linkcolor=darkblue, urlcolor=darkblue}

\usepackage{wrapfig}
\usepackage{spverbatim}
\usepackage{tikz}
\usetikzlibrary{shapes.geometric}
\newcommand{\warningsign}{\tikz[baseline=-.75ex] \node[shape=regular polygon, regular polygon sides=3, inner sep=0pt, draw, thick] {\textbf{!}};}

\AtBeginDocument{%
\providecommand\BibTeX{{%
\normalfont B\kern-0.5em{\scshape i\kern-0.25em b}\kern-0.8em\TeX}}}

\newcommand{\tableskip}{\noalign{\vskip 2pt}}

\newcommand{\sribd}{
\hspace{-5pt}
  $^{\includegraphics[width=0.3cm]{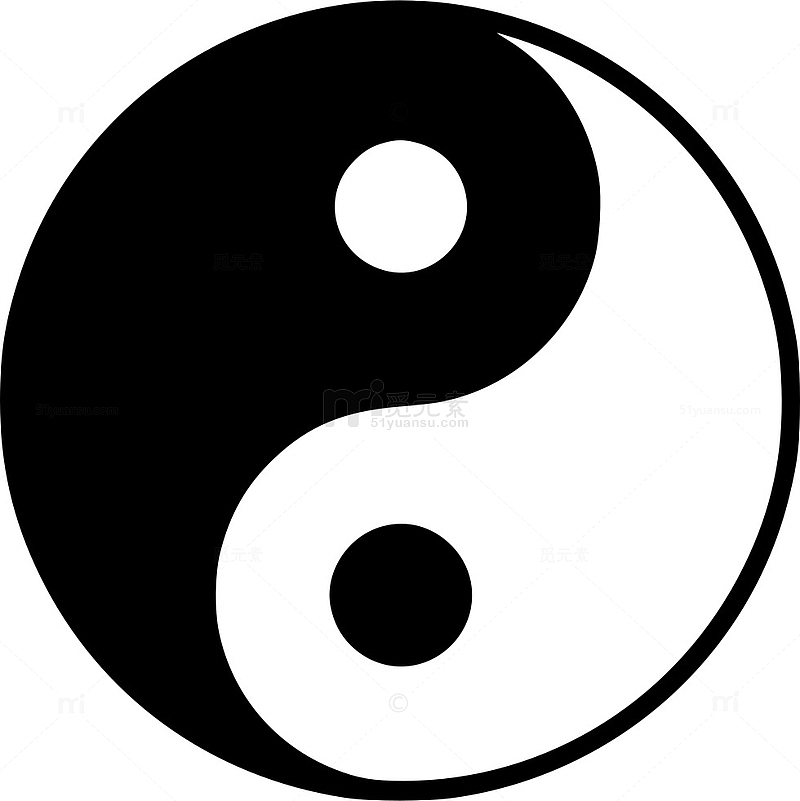}}$
\hspace{-5pt}
}

\newcommand{\BIGSRIBD}{
\includegraphics[width=0.3cm]{pic/01.jpg}
}

\newcommand{\cuhksz}{
\hspace{-5pt}
  $^{\includegraphics[width=0.35cm]{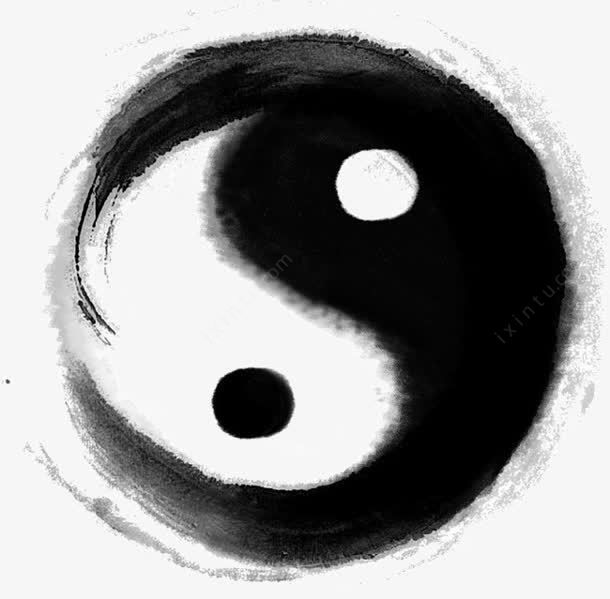}}$
\hspace{-5pt}
}

\newcommand{\BIGCUHKSZ}{
  {\includegraphics[width=0.35cm]{pic/10.jpg}}
}

\newcommand{\both}{
\hspace{-5pt}
  $^{\includegraphics[width=0.35cm]{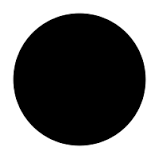}}$
\hspace{-5pt}
}

\usepackage{multirow}
\usepackage{tabularx}
\usepackage{makecell}
\usepackage{CJK}
\usepackage{geometry}
\usepackage{booktabs}
\usepackage{color} 
\usepackage{amssymb} 
\usepackage{pifont}
\definecolor{lightgray}{gray}{0.9}

\usepackage[most]{tcolorbox}
\usepackage{float}
\usepackage{xspace}
\tcbset{
aibox/.style={
width=430pt,
top=10pt,
colback=white,
colframe=black,
colbacktitle=black,
enhanced,
center,
attach boxed title to top left={yshift=-0.12in,xshift=0.15in},
boxed title style={boxrule=0pt,colframe=white},
}
}
\newtcolorbox{AIbox}[2][]{aibox,title=#2,#1}

\newcommand{\dec}[1]{{\small\hspace{0.05cm}{\color[HTML]{CD5C5C}{$_{\textbf{-#1}}$}}}}

\newcommand{\imp}[1]{{\small\hspace{0.05cm}{\color[HTML]{32CB00}{$_{\textbf{+#1}}$}}}}

\usepackage{tabularx}

\usepackage{times}
\usepackage{latexsym}

\usepackage[T1]{fontenc}

\usepackage[utf8]{inputenc}

\usepackage{inconsolata}

\title{Apollo: A Lightweight Multilingual Medical LLM towards Democratizing Medical AI to 6B People}

\author{Xidong Wang\cuhksz$^\dagger$,~Nuo Chen\cuhksz$^\dagger$, Junying Chen\both$^\dagger$,~Yidong Wang\cuhksz,~Guorui Zhen\cuhksz, Chunxian Zhang\cuhksz\\
\textbf{Xiangbo Wu}\cuhksz,~\textbf{Yan Hu}\cuhksz,~\textbf{Anningzhe Gao\sribd,~Xiang Wan\sribd,~Haizhou Li\both,~Benyou Wang\both}\thanks{ Benyou is the corresponding author (A\textit{wangbenyou@cuhk.edu.cn}); first three authors contributed to this work equally.  The democratization of Medical AI involves making Medical AI technologies more accessible, especially in areas without native, open-source LLMs, and providing streamlined versions for those with limited resources.
} \\
\BIGCUHKSZ The Chinese University of Hong Kong, Shenzhen\\
\BIGSRIBD Shenzhen Research Institute of Big Data\\
\url{https://github.com/FreedomIntelligence/Apollo} \\
\url{https://apollo.llmzoo.com/}\\
}

\begin{document}
\maketitle

\begin{abstract}
Despite the vast repository of global medical knowledge predominantly being in English, local languages are crucial for delivering tailored healthcare services, particularly in areas with limited medical resources. To  extend the reach of medical AI advancements to a broader population, we  aim to develop medical LLMs across the six most widely spoken languages, encompassing a global population of 6.1 billion. 
This effort culminates in the creation of the \textbf{ApolloCorpora}  multilingual  medical dataset and the \textbf{XMedBench} benchmark. In the multilingual  medical benchmark, the released \textbf{Apollo} models, at various relatively-small sizes (i.e., 0.5B, 1.8B, 2B, 6B, and 7B), achieve the best performance  among models of equivalent size. Especially, Apollo-7B is the state-of-the-art multilingual medical LLMs up to 70B. Additionally, these lite models could be used to improve the multi-lingual medical capabilities of larger models \textbf{without fine-tuning} in a proxy-tuning fashion.  We will open-source training corpora, code, model weights and evaluation benchmark.
\end{abstract}
\section{Introduction}


The integration of medical knowledge and artificial intelligence has always been a focal point of research communities, with each incremental improvement potentially enhancing patient experiences and healing rates—serving as a direct manifestation of technology for good. Although medical large language models are promising, existing works are mainly in Chinese~\citep{chen2023huatuogpt, zhang2023huatuogpt, bao2023disc} or English~\citep{wu2023pmc, chen2023meditron}.
The multilingual adaption of medical LLMs, as part of the democratization of large models,  seeks to extend the benefits of cutting-edge LLMs to a broader spectrum of users, including those from underrepresented communities. This movement is akin to the historical endeavors to disseminate transformative technologies like electricity and vaccines to wider communities, positing LLMs as the modern equivalents of these essential innovations.


\begin{figure*}[!h]
\centering
\includegraphics[width=0.9\linewidth]{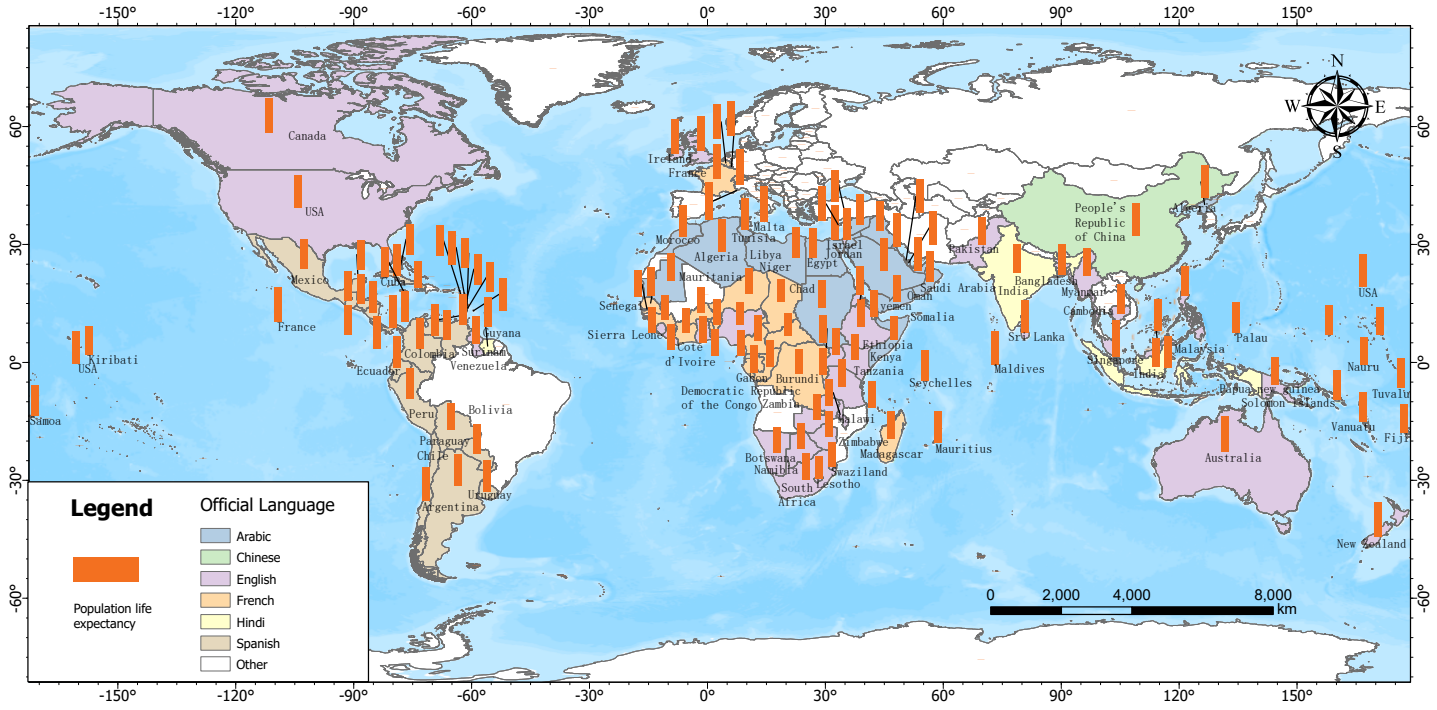}
\caption{Countries covered by ApolloCorpora and relative population life expectancy}
\label{fig:map}
\end{figure*}

\begin{figure*}[h]
\centering
\includegraphics[width=0.9\linewidth]{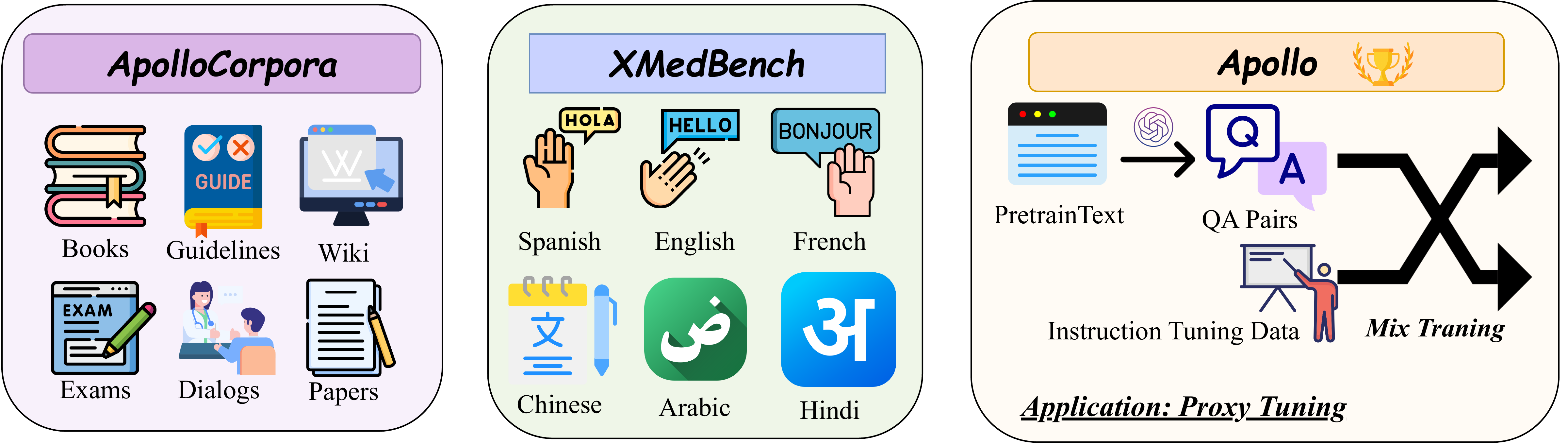}
\caption{Overview of this work, including corpora, benchmark, models and their application.}
\label{fig:oveview}
\end{figure*}

\paragraph{Rationale of Multilinguality in Medical LLMs}
The rationale of multilingual  adaption in medical LLMs~\citep{cox2021multilingual, pecina2014adaptation} could be twofold.
Firstly, non-native English-speaking doctors often engage in bilingual learning through their native language and English from the outset, naturally introducing multilingual challenges in the learning process~\citep{marko2006towards}. Secondly, to better serve local communities, especially in countries and regions with scarce medical resources, medical aid based on local languages often achieves higher communication efficiency and acceptance~\citep{brindley2014improving, albrecht2013usage}. Meanwhile, local medical knowledge can complement mainstream medical knowledge, fostering mutual benefits to accelerate medical development~\citep{klayman1985qinghaosu, yuan2016traditional}. Our pilot study in Sec.~\ref{sec:pilot} also reveals that \textit{joint training of multiple languages enhances performance in the medical LLMs}, indicating a beneficial complementarity among languages. 

\paragraph{The  Corpora: ApolloCorpora}
Towards building multilingual medical LLMs, the first step is to build high-qaulity corpora. We select the six most populous languages: English, Chinese, Hindi, Spanish, French, and Arabic for experiments \footnote{See the language popularity in \url{https://en.wikipedia. org/wiki/List_of_languages_by_total_number_of_speakers}, The set of the selected six languages covers a total of 6.1 billion people in 132 countries and regions, according to~\url{https://en.wikipedia.org/wiki/List_of_countries_and_dependencies_by_population}.}.  
As shown in Fig. \ref{fig:map}, it encompasses a diverse array of linguistic backgrounds, particularly in regions often characterized by limited medical resources (e.g., areas with lower average life expectancies).
We collect and process data from data sources including books, clinical guidelines, encyclopedias, papers, online forums and examinations, obtaining \textbf{ApolloCorpora} with 2.5B tokens.


%


\paragraph{The Lite LLM: Apollo}
The resulted  multilingual medical LLMs trained by ApolloCorpora  are named \textbf{Apollo}, 
to commemorate the Greek deity associated with healing, disease, the Sun, and light; this symbolizes the democratization of medical LLMs to 6 billion people, illuminating global healthcare.
We explore a new domain adaption method that rewrite the pre-training corpora into QA pairs using ChatGPT~\citep{li2023self} and adaptively sample training data, resulting in a smoother transition compared with the typical paradigm with continued pretraining and instruction tuning.
Apollo ranges from 2B to 7B parameters.
The advantage of the relatively-small model scale  includes potential use as draft models for speculative decoding~\citep{leviathan2023fast} or as proxy models for proxy-tuning~\citep{liu2024tuning}. In particular, we apply proxy-tuning on top of Apollo to larger \textit{general} LLMs, enhancing its \textit{multilingual medical} capabilities. This is achieved without the need to directly train the \textit{general} model using sensitive medical corpora, thereby underscoring the practical significance of Apollo in terms of protecting the privacy of medical training data against centralized training methods.

\paragraph{The Benchmark: XMedBench}
We  select local multiple-choice tasks to assess models' medical knowledge. For Hindi and Arabic, which lack local assessments, we choose to translate the medical-related parts of MMLU~\citep{hendrycks2020measuring}. The results show that the gap between open source and closed source is narrowing. 
While GPT-4 has demonstrated superior efficacy across numerous languages, the Apollo series models achieve the best performance among models of equivalent size.


\paragraph{Contributions}
As shown in Fig.~\ref{fig:oveview}, the contributions of this paper are as follows. (1) We collect and organize a high-quality multilingual medical corpora \textbf{ApolloCorpora} with rich local characteristics; (2) We obtain a series of \textbf{SOTA} multilingual LLMs \textbf{Apollo} at various parameter scales (especially in relatively-small sizes)
; (3) By using proxy-tuning,  Apollo could significantly improve larger general LLMs \textbf{without finetuning}, providing a new way to mitigating the exposure of private medical  training data to centralized training systems; (4) We introduce a  multilingual medical evaluation \textbf{XMedBench} and conduct extensive benchmark of existing models. 



\section{Pilot Study on the Multilinguality  of Medical LLMs}
\label{sec:pilot}
%
\subsection{The Research Question}
This section presents two contrasting hypotheses regarding the nature of medical knowledge and its representation in LLMs.

\paragraph{Language-neutral Hypothesis}
It is commonly believed that knowledge, whether medical or general, should be independent of language. For example, the fact that \textit{the sun rises in the east} remains unchanged whether expressed in English or Chinese, suggesting that knowledge might be considered \textit{language-neutral}. Consequently, medical corpora in various languages could serve as an augmentation for training, thereby enhancing the efficacy of the resulting medical LLMs.

\paragraph{Language-dependent Hypothesis}
However, due to historical, cultural, and regional political influences, medical knowledge can vary significantly across different cultural contexts, especially as reflected in language. 
The integration of medical knowledge across languages  might dilute the local specificity of medicine due to differences in lifestyle and constitution across regions~\citep{rotti2014determinants, sharma2020group}.
For instance, in traditional Chinese medicine, colds are classified into types caused by heat or cold, and treatments may vary locally, relying on unique medical recipe that uses local herbs or substances. This indicates that historical and cultural traditions can shape medical knowledge to some extent. 

Consequently, there arises a \textbf{research question} in the context of medical LLMs: 
\begin{quote}
\textit{Can medical data in different languages complement or harm each? }
\end{quote}
This leads us to explore whether medical corpora in different languages supplement each other or conflict within medical LLMs.

\subsection{The Pilot Study}

\paragraph{Experimental settings} To investigate the above question, we use a lite  multilingual LLM Qwen-1.8B~\citep{bai2023qwen} as the LM backbone. It is argued that the findings should be agnostic to the selection of LM backbones, the selection of LLM Qwen-1.8B is due to its popularity, performance and more the importantly multilingual support. In the
 \textbf{monolingual training} setting, the LM backbone is further trained by individual language, resulted six language-specific LLM variants (i.e.
    , English, Chinese, French, Spanish, Arabic, and Hindi.). The training data used could be found in Sec.~\ref{sec:data}. Moreover, we average the weights of these six LLM variants and obtain a new model that does not need further training, denoted as `\textbf{weight average}'.
 In the \textbf{multilingual training} setting, we train the LM using a mixture of the corpora in these six languages.

\begin{table*}[h]\footnotesize
\centering
\begin{tabular}{l|cccccc|c}
\hline
\textbf{Model} & \textbf{English} & \textbf{Chinese} & \textbf{French} & \textbf{Spanish} & \textbf{Arabic} & \textbf{Hindi} & \textbf{Avg.} \\ \hline
\rowcolor{gray!15}\multicolumn{8}{c}{\textbf{Base Model}} \\ \tableskip 
Qwen-1.8B & 32.91  & 40.07  & 22.12 & 27.43 & 23.71 & 8.82 & 25.84 \\
\hline \rowcolor{gray!15}\multicolumn{8}{c}{\textbf{Language Specific Models}} \\ \tableskip
Apollo-\textit{English} & \cellcolor{gray!25}39.44  & 45.27 & 28.35 & 31.76 & 22.61 & 8.72 & 29.36 \\
Apollo-\textit{Chinese}& 39.42 & \cellcolor{gray!25}61.13 & 28.97 & 33.83 & 27.94 & 25.34 & 36.11 \\
Apollo-\textit{French} & 30.94 & 32.71 & \cellcolor{gray!25}23.81 & 27.00 & 24.54 & 1.74 & 23.46 \\
Apollo-\textit{Spanish}& 33.84 & 43.81 & 27.41 & \cellcolor{gray!25}35.39 & 28.40 & 23.88 & 32.12 \\
Apollo-\textit{Arabic} & 36.40 & 44.27 & 3.74 & 15.73 & \cellcolor{gray!25}25.90 & 3.03 & 21.85 \\
Apollo-\textit{Hindi} & 25.18 & 3.45 & 18.38 & 19.69 & 1.00 & \cellcolor{gray!25}25.53 & 15.54 \\ 
\hline \rowcolor{gray!15}\multicolumn{8}{c}{\textbf{Our Method}} \\ \tableskip
Apollo{\footnotesize~(weight average)} & 40.54  & 45.58 & 28.04 & 34.08 & 28.95 & 24.06 & 33.54 \\
Apollo{\footnotesize~(multilingual training)} & \textbf{45.43}  & \textbf{62.93} & \textbf{38.01} & \textbf{42.15} & \textbf{34.74} & \textbf{25.62} & \textbf{41.48} \\
\hline
\end{tabular}
\caption{The pilot study on monolingual training and multilingual training. It shows the average accuracy among datasets in each language, see  details in Sec.~\ref{sec:benchmark}.}

\label{tab:pilot_study}
\end{table*}

\begin{table*}[h]\footnotesize
\resizebox{\textwidth}{!}{
\begin{tabular}{@{}llll@{}}
\toprule
Data Source & Training Stage& Language (\# Token) & \# Token \\ \midrule
\multicolumn{4}{l}{\textit{High-quality medical data}} \\
Books & Continue Pretrain & EN (296.7M), ZH (117.1M)& 413.8M \\
Papers& Continue Pretrain & ZH (45.6M), EN (252.9M), ES (46.0M), FR (4.5M)& 349.0M \\
Encyclopedias & Continue Pre-train & EN (221.1M), FR (4.6M), HI (0.5M) & 226.2M \\
Dialogues & Continue Pretrain & EN (92.1M), ZH (46.6M), AR (10.4M)& 149.1M \\
Exams & Instruction Tuning & EN (42.1M), ZH (35.3M), FR (0.1M), ES (0.5M)& 78.0M\\
Guidelines& Continue Pretrain & EN (29.6M)& 29.6M\\ \midrule
\multicolumn{4}{l}{\textit{Data entry outside the profession}}\\
Web & Continue Pretrain & EN (499.9M), ZH (329.3M), ES (57.5M)& 886.7M \\
General & Instruction Tuning & EN (194.5M), ZH (69.4M), HI (43.9M), FR (20.0M), AR (18.7M), ES (18.4M) & 364.9M \\
Math& Instruction Tuning& EN (18.9M), ZH (3.7M) & 22.6M\\
Code& Instruction Tuning& EN (9.2M), ZH (7.2M)& 16.4M\\ \bottomrule
\end{tabular}%
}
\caption{Taxonomy of ApolloCorpora and Token statistics}
\label{tab:data_tax}
\end{table*}

\paragraph{Findings}
Tab.~\ref{tab:pilot_study} highlights the effectiveness of our methods in leveraging multilingual data to enhance the performance of medical Large Language Models (LLMs).  The language-specific models, each trained exclusively on data from one language, demonstrate varying degrees of improvement in their respective languages over the original LM, underscoring the value of language-specific training. However, these models show limitations outside their target languages, as seen in the relatively low scores in non-target languages, particularly in the Apollo-French-1.8B and Apollo-Hindi-1.8B models. Our method, which includes both the \textbf{weight average} and m\textbf{multilingual training}, significantly outperforms language-specific models across all languages in terms of average performance, as shown in the last column of Tab. \ref{tab:pilot_study}. This illustrates the substantial benefits of combining multilingual data for training medical LLMs, with marked improvements in understanding and generating medical information across a diverse set of languages.

Therefore, we conclude the finding as below.
\begin{quote}
\textit{In general,  multilingual medical corpora benefits medical LLMs. }
\end{quote}


{\color{red}
\warningsign} \textbf{Potential Risks of  Multilingual training in medical LLMs.}
While acknowledging the potential for conflicts arising from integrating language-specific medical knowledge in multilingual training, we recognize this as a risk inherent in such an approach. However, based on the average performance improvements observed in Tab.~\ref{tab:pilot_study}, we are inclined to believe in the efficacy of multilingual training, especially in the context of medical knowledge, which we argue to be language-neutral to a significant extent. We propose that the conflicts or the potential undermining of local specificities observed in multilingual training be considered as an area for future research. This perspective invites further exploration into how multilingual LLMs can be optimized to respect and preserve the unique medical practices and knowledge embedded within each language, while still harnessing the collective benefits of a multilingual approach.

\section{Corpora and Model Training for Apollo}
\label{sec:data}





\subsection{ApolloCorpora: Data Collection and Cleaning}
\label{sec:phi}



After extensive communication with doctors and medical students, we identified six high-quality medical data collections: medical books, medical encyclopedias, medical clinical guidelines, medical papers, medical examinations, and professional doctor-patient dialogues, see Tab.~\ref{tab:data_tax}. 
To mimic the diverse learning experience of medical students beyond their core professional studies, we also included a wide range of medical-related content from the Internet. This approach captures the evolving nature of medical information found online. Additionally, we incorporated tasks that require mathematical reasoning and coding. This inclusion enriches the model's skill set with critical analytical and problem-solving abilities, essential for the multifaceted demands of medical practice.

Regarding the \textbf{License issue}, we only screen data sets with complete open source protocols during the collection process to ensure that the open source protocols are friendly while ensuring quality. 
Inspired by~\citep{cheng2023adapting,chen2023huatuogpt}, we use ChatGPT~\footnote{gpt-3.5-turbo-16k-0613} to generate questions and answers for a certain paragraph. For paragraph interception, we divide it according to the basic semantic units in the data set. 
Regarding the \textbf{quality assurance}, we rely on the help of doctors to carefully control the quality from the source of the data. 
The details of data collection and prepossessing are shown in App.~\ref{sec:ApolloCorpora}.



For the dimension of \textbf{multilingual medical expertise}, we insist on using only medical data sets entirely from local languages and do not translate any medical-related data. This is done out of the following two considerations. First, there are many related works that prove that medical translation is a very complex task that cannot be simply solved by translation software; second, the expression habits of different languages, effective drugs, and even culture and Taboo terms arising from faith need to come from the local community intact, so as to maximize communication efficiency and avoid conflicts.  See localized features in App.~\ref{sec:localization}.

\paragraph{Data Leakage Checking}
The issue of data leakage is a recent focus of the academic community, which largely determines whether the results of the paper are convincing. For knowledge embedding tasks, data leakage screening with different stringency often leads to different performance. We follow Med-PaLM2~\citep{singhal2023towards} and adopt a more stringent deletion strategy. Specifically, we define a data item as leaked data if the entire question or at least 64 consecutive characters overlap with the data item. Regarding the exam exercise data source, there were 580,645 exercises before screening, and 3,041 exercises are deleted, with a screening rate of 0.52\%. For other data sources, since they are not exam questions, there is no difference before and after filtering.

\subsection{Apollo, the Lite Multilingual Medical LLM}
We have two main starting points for training small models. First, medical equipment usually cannot call network services due to its strict privacy protection settings. For local services, the small model can achieve offline inference on the PC side, ensuring complete data localization to help improve the efficiency of medical staff; secondly, the original intention of our article is to explore a reproducible technical solution at an affordable computing cost, and promote the exploration of the field and the raising of new questions. Small models are useful for The training is very friendly for academic researchers who lack sufficient computing power.


Training models in the medical field usually involves continuing pre-training on the corpus. However, some scholars believe that although training on the original corpus gives the model domain knowledge, it greatly damages its ability to prompt question answers~\citep{cheng2023adapting}. We consider exploring ways to rewrite the pre-training corpus into the form of question-and-answer pairs to alleviate this problem~\citep{chen2023huatuogpt}. At the same time, we use priority sampling methods to achieve a smooth transition between continued pre-training and Instrcution Tuning to ensure the continuity of learning rate and data distribution transformation.

\paragraph{Mix Training} Our dataset $D$ comprises continuing pre-training data $D_{PT}$ and instruction tuning data $D_{SFT}$. The sampling probability of each data $x \in D$ changes during training. The sampling probability of data $x$ at step $t$ during training was determined using priority sampling, defined as:
$$P_t(x) = \frac{\pi(x)}{\sum_{y \in D-S_t} \pi(y)}$$
Here, $\pi(x)$ denotes the priority of element $x$, and $S_t$ represents the sampled data before step $t$. 





\begin{figure*}[h]\footnotesize
\begin{AIbox}{Prompt}
User:You are a medical doctor answering real-world medical exam questions. Select one correct answer from A to D. \space\space\space\space  Question: \{{\tt question}\}\\
Options: (A) \{{\tt option\_a}\} (B) \{{\tt option\_b}\} (C) \{{\tt option\_c}\} (D) \{{\tt option\_d}\} \\
Assistant:The correct answer is \{answer\}. \textless special\_token\ \textgreater
\end{AIbox} 
\caption{Prompt Template for XMedBench}
\label{fig:prompt} 
\end{figure*}

\begin{table*}[h]
\centering
\resizebox{1.\textwidth}{!}{
\setlength\tabcolsep{2.5pt}

\begin{tabular}{l|ccc|cc|cccc|l}
\hline
\multirow{2}{*}{\textbf{Language}} & \multicolumn{3}{c|}{\textbf{English}} & \multicolumn{2}{c|}{\textbf{Chinese}} & \textbf{French} & \textbf{Spanish} & \textbf{Arabic} & \textbf{Hindi} & \multirow{2}{*}{\textbf{Avg.}} \\ 
& \textbf{USMLE} & \textbf{MedMCQA} & \textbf{MMLU$\diamondsuit$} & \textbf{MCMLE} & \textbf{CMMLU$\diamondsuit$} & \textbf{FrenchMedMCQA} & \textbf{HEAD-QA} & \textbf{MMLU$\diamondsuit$} & \textbf{MMLU$\diamondsuit$} &  \\
\hline \rowcolor{gray!15}\multicolumn{11}{c}{\textbf{Closed-source}} \\ \tableskip
GPT-4 & 79.10 & 70.40 & 86.00 & 65.72 & 65.72 & 89.72 & 85.05 & 56.43 & 62.17 & 73.37 \\ 
GPT-3.5 & 61.98 & 56.51 & 72.94 & 58.73 & 50.41 & 68.54 & 71.48 & 39.70 & 39.94 & 57.80 \\
\hline
\rowcolor{gray!15}\multicolumn{11}{c}{\textbf{Open-source (Above 70B)}} \\ \tableskip
Qwen-72B & 64.10 & 62.16 & 78.46 & 91.68 & 81.47 & 74.14 & 76.62 & 46.87 & 43.16 & 68.74 \\
Meditron-70B & 55.70 & 50.87 & 69.59 & 48.34 & 40.29 & 53.27 & 59.74 & 19.30 & 31.31 & 47.60 \\

Llama-2-70B & 32.99 & 48.29 & 64.62 & 25.80 & 25.13 & 50.47 & 54.34 & 1.65 & 26.35 & 36.63 \\
\hline
\rowcolor{gray!15}\multicolumn{11}{c}{\textbf{Open-source (Above 7B)}} \\ \tableskip
Qwen-14B & 50.27 & 45.83 & 61.68 & 75.22 & 61.82 & 49.53 & 60.81 & 36.58 & 32.29 & 52.67 \\
Gemma-7B& 53.42& 50.94& 70.15& 48.95& 43.29& 57.63& 62.79& 36.21& 48.58& 52.44 \\
MMedLM2-7B & 55.46 & 50.49 & 68.15 & 64.30 & 56.11 & 58.57 & 62.14 & 23.53 & 24.15 & 51.45\\
Yi-34B & 62.45 & 60.60 & 71.86 & 26.12 & 26.51 & 66.04 & 69.99 & 30.70 & 9.73 & 47.00 \\
Mistral-7B & 47.29 & 47.38 & 62.80 & 38.32 & 34.21 & 50.78 & 51.93 & 28.40 & 27.36 & 43.16 \\
Qwen-7B & 32.36 & 39.52 & 53.22 & 54.32 & 44.71 & 37.69 & 45.05 & 28.31 & 24.89 & 40.01 \\
Zephyr-7B-$\beta$ & 41.95 & 42.48 & 58.74 & 36.11 & 31.88 & 46.42 & 46.77 & 27.02 & 27.92 & 39.92 \\
BioMistral-7B & 41.79 & 42.05 & 54.46 & 34.65 & 31.43 & 43.61 & 44.66 & 27.11 & 22.96 & 38.08 \\
Huatuo2-7B & 37.86 & 36.58 & 42.49 & 55.08 & 43.81 & 27.41 & 33.88 & 25.92 & 27.46 & 36.72 \\

Huatuo2-13B & 29.77 & 36.58 & 42.86 & 56.07 & 45.46 & 22.42 & 36.13 & 18.29 & 13.59 & 33.46 \\
Llama-2-7B & 32.13 & 36.58 & 40.14 & 25.39 & 25.13 & 29.60 & 33.54 & 21.42 & 27.27 & 30.13 \\
Meditron-7B & 33.78 & 34.54 & 36.18 & 27.50 & 27.16 & 24.00 & 32.81 & 1.65 & 18.27 & 26.21 \\
PMC-Llama-7B & 20.11 & 23.12 & 19.72 & 16.90 & 16.73 & 17.13 & 18.68 & 9.65 & 2.85 & 16.10 \\
\hline \rowcolor{gray!15}\multicolumn{11}{c}{\textbf{Our Models}} \\ \tableskip
\textbf{Apollo-0.5B} & 32.99 & 37.82 & 45.87 & 56.57 & 42.08 & 27.41 & 36.67 & 31.89 & 25.90 & 37.47 \\
\textbf{Apollo-1.8B} & 42.18 & 44.99 & 49.12 & 72.30 & 53.56 & 38.01 & 42.15 & 34.74 & 25.62 & 44.74 \\
\textbf{Apollo-2B}& 38.33 & 42.00 & 52.89 & 46.76 & 36.76 & 38.32 & 41.28 & 31.62 & 31.50 & 39.94 \\
\textbf{Apollo-6B} & 56.25 & 57.53 & 68.65 & 85.52 & 72.62 & 51.71 & 58.47 & 33.46 & 33.61 & 57.54 \\

\textbf{Apollo-7B} & 56.00 & 58.21 & 71.86 & 72.36 & 59.04 & 60.44 & 63.73 & 41.82 & 45.55 & \textbf{58.78} \\
\hline
\end{tabular}
}
\caption{Performance comparison across various medical question answering models.}
\label{tab:benchmark}
\end{table*}

 \paragraph{Settings} We set the priority $\pi(x)=16$ for $x \in D_{PT}$, and $\pi(x)=2$ for $x \in D_{SFT}$. In order to achieve the purpose of smooth transition of sampling ratio. The overall training sequence of pre-training corpus first, and then instruction Tuning corpus is maintained, but the transition can be smoothed. The Batch size of model training is set to 256, the learning rate is set to 1e-5, and the warm up rate of Cosine scheduler is set to 0.03. The pre-training corpus is trained for one epoch, the instruction data is trained for two epochs.




\section{Evaluation}

\subsection{XMedBench: Multilingual Medical Knowledge Evaluation}
\label{sec:benchmark}


We focus on assessing multilingual medical knowledge, select multiple-choice questions as tasks, and collect common data sets with local medical characteristics, see details in App.~\ref{sec:bench}.

\paragraph{Construction of XMedBench}
\label{sec:bench_construct}
For English, we use the MedQA-USMLE~\citep{zhang2018medical}, MedMCQA~\citep{pal2022medmcqa}, and medical-related parts of MMLU~\citep{hendrycks2020measuring}; for Chinese, we used the of MedQA-MCMLE~\citep{zhang2018medical} and medical-related parts CMMLU~\citep{li2023cmmlu}; for Spanish, we used HEAD-QA~\citep{vilares2019head}; for French, we used FrenMedMCQA~\citep{labrak2023frenchmedmcqa}; For Arabic and Hindi, which lack local evaluation, we compromised and followed Llama3's multilingual evaluation method~\citep{dubey2024llama3herdmodels}, using Google Translate and inviting practicing physicians to proofread and finally get the translated version of MMLU. Specifically, we follow Med-PaLM2~\citep{singhal2023towards} and select six subcategories in MMLU: Clinical knowledge, Medical genetics, Anatomy, Professional medicine, College biology, and College medicine. For MedQA, we choose the 4-options version. For CMMLU, we select seven subdirectories: Anatomy, Clinical knowledge, College medicine, Genetics, Nutrition, Traditional chinese medicine, and Virology.



 \paragraph{Settings}
We adopt 3-shot evaluation and use regular matching to extract options. The specific evaluation prompts are shown in Fig. \ref{fig:prompt}. For the generation strategy, we do not perform sampling and set the maximum and minimum number of generated tokens to 128 and 2. For model loading, except for the 0.5B size model which uses full precision loading, we uniformly use half precision loading. Please see the App. \ref{sec:bench_models} for details of Models.

\begin{table*}[]\footnotesize
\centering
\setlength\tabcolsep{9pt}
\vspace{1mm}
\begin{tabular}{lccccccc}
\hline
\textbf{Model} & \textbf{English} & \textbf{Chinese} & \textbf{French} & \textbf{Spanish} & \textbf{Arabic} & \textbf{Hindi} & \textbf{Avg.} \\ \hline
\rowcolor{gray!15}\multicolumn{8}{c}{\textbf{Base Model}} \\ \tableskip
Qwen-1.8B & 32.91  & 40.07 & 22.12 & 27.43 & 23.71 & 8.82 & 25.84 \\
\hline \rowcolor{gray!15} \multicolumn{8}{c}{\textbf{Rewrite Pre-training Data into QA}} \\ \tableskip
ParaData-Sep-1.8B & \textbf{47.34} & 57.58 & 37.69 & 41.24 & 29.32 & 15.79 & 38.16 \\
QAData-Sep-1.8B & 45.43  & 59.21 & 38.01 & \textbf{42.48} & 31.43 & 14.60 & 38.53 \\
\hline \rowcolor{gray!15} \multicolumn{8}{c}{\textbf{Smoothly Transition the Two Stages}} \\ \tableskip
ParaData-Mix-1.8B & 42.97  & 53.56 & 33.02 & 36.88 & 31.71 & 14.23 & 35.40 \\
\textbf{QAData-Mix-1.8B (Apollo-1.8B)} & 45.43 & \textbf{62.93} & \textbf{38.01} & 42.15 & \textbf{34.74} & \textbf{25.62} & \textbf{41.48} \\
\hline
\end{tabular}
\caption{Mix Training for Multilingual.}
\label{tab:mix_performance}
\end{table*}

\begin{table*}[]\footnotesize
\centering

\resizebox{\textwidth}{!}{
\setlength\tabcolsep{2.5pt}
\begin{tabular}{l|ccc|cc|cccc|l}
\hline
\multirow{2}{*}{\textbf{Language}} & \multicolumn{3}{c|}{\textbf{English}} & \multicolumn{2}{c|}{\textbf{Chinese}} & \textbf{French} & \textbf{Spanish} & \textbf{Arabic} & \textbf{Hindi} & \multirow{2}{*}{\textbf{Avg.}} \\ 
& \textbf{USMLE} & \textbf{MedMCQA} & \textbf{MMLU$\diamondsuit$} & \textbf{MCMLE} & \textbf{CMMLU$\diamondsuit$} & \textbf{FrenchMedMCQA} & \textbf{HEAD-QA} & \textbf{MMLU$\diamondsuit$} & \textbf{MMLU$\diamondsuit$} &  \\
\hline \rowcolor{gray!15}\multicolumn{11}{c}{\textbf{Our Models and their Bases}} \\ \tableskip
Qwen-0.5B & 24.43 & 3.78 & 16.94 & 14.16 & 10.88 & 23.68 & 26.02 & 26.29 & 26.35 & 19.17 \\
\textbf{Apollo-0.5B} & 32.99 & 37.82 & 45.87 & 56.57 & 42.08 & 27.41 & 36.67 & 31.89 & 25.90 & 37.47 \\
Qwen-1.8B & 26.79 & 31.05 & 40.89 & 44.28 & 35.86 & 22.12 & 27.43 & 23.71 & 8.82 & 28.99 \\
\textbf{Apollo-1.8B} & 42.18 & 44.99 & 49.12 & 72.30 & 53.56 & 38.01 & 42.15 & 34.74 & 25.62 & 44.74 \\
Gemma-2B& 30.24& 32.27 & 37.35 & 25.98 & 28.06 & 25.86 & 32.43 & 20.96 & 25.53 & 28.74 \\
\textbf{Apollo-2B}& 38.33 & 42.00 & 52.89 & 46.76 & 36.76 & 38.32 & 41.28 & 31.62 & 31.50 & 39.94 \\
Yi-6B & 45.48 & 47.98 & 62.27 & 78.90 & 69.47 & 45.79 & 47.01 & 12.22 & 10.74 & 46.65 \\
\textbf{Apollo-6B} & 56.25 & 57.53 & 68.65 & 85.52 & 72.62 & 51.71 & 58.47 & 33.46 & 33.61 & 57.54 \\
Gemma-7B& 53.42& 50.94& 70.15& 48.95& 43.29& 57.63& 62.79& 36.21& 48.58& 52.44 \\
\textbf{Apollo-7B} & 56.00 & 58.21 & 71.86 & 72.36 & 59.04 & 60.44 & 63.73 & 41.82 & 45.55 & \textbf{58.78} \\

\hline
\end{tabular}}
\caption{Model performance comparison before and after Mix Training}
\vspace{2mm}
\label{tab:bases}
\end{table*}

\subsection{Benchmarking results}
As shown in Tab. \ref{tab:benchmark}, GPT-4 and Qwen-72B rank first in closed source and open source with accuracy rates of 73.37 and 68.74 respectively. The gap between closed source and open source is decreasing. The Apollo series models achieve the best performance of models of the same size. Apollo-7B achieve comparable performance as GPT-3.5, Apollo-1.8B achieve comparable performance as Mistral-7B, and Apollo-0.5B achieve comparable performance as Llama2-7B.

From a \textbf{language} perspective, all models scored worse on Arabic and Hindi compared to English, which further demonstrates the medical community’s neglect of these two languages. Note that GPT-4 support these languages better, reflecting OpenAI’s emphasis on multi-language scenarios. Mistral is better adapted to French, and the Qwen and Yi models have better support for Chinese.

\subsection{More Analysis}

As shown in the Tab. \ref{tab:mix_performance}, under the experimental setting of pre-training first and then SFT, rewriting the pre-training into question and answer has no loss on the overall effect of the model. We also notice that the performance drop after rewriting in English and Hindi, but other languages' performance improve.
After adopting the smooth transition method, we find that except English, other languages' performance greatly improve. This may be because the data distribution transformation of the previous method is too rigid, resulting in the inability to learn knowledge of long-tail languages (such as Hindi). Using the method of converting to question and answer pairs and making a smooth transition may be able to minimize the knowledge loss of distribution transformation, allowing the model to fully learn the knowledge of long-tail languages and improve the ability of non-mainstream languages. 

As shown in the Tab. \ref{tab:bases}, models' multilingual medical capabilities have been significantly improved after Mix Training. For different model sizes, although the improvement effect gradually decreases as the model parameters increase, the model performance still continues to increase, indicating promising prospects for scaling up training. Impressively, Apollo-0.5B achieves considerable performance with few parameters. Given its potential for real-time inference on a wide range of hardware, we believe it can democratize advances in medical AI to the broader community.

\begin{table*}[h]\footnotesize
\centering
\setlength\tabcolsep{5.5pt}
\begin{tabular}{l|llllll|l}
\hline
\textbf{Model} & \textbf{English} & \textbf{Chinese} & \textbf{French} & \textbf{Spanish} & \textbf{Arabic} & \textbf{Hindi} & \textbf{Avg.} \\ 
 \hline \rowcolor{gray!15}\multicolumn{8}{c}{\textbf{Other Models}} \\ \tableskip
GPT-3.5 & 63.81  & 54.57 & 68.54 & 71.48 & 39.70 & 39.94 & 56.34 \\
Meditron-7B & 34.83  & 27.33 & 24.00 & 32.81 & 1.65 & 18.27 & 23.15 \\
\hline \rowcolor{gray!15}\multicolumn{8}{c}{\textbf{Proxy-Tuning for Qwen}} \\ \tableskip
Apollo-1.8B (from Qwen-1.8B) & 45.43  & 72.30 & 38.01 & 42.15 & 34.74 & 25.62 & 43.04 \\
Qwen-7B & 41.70 & 49.52 & 37.69 & 45.05 & 28.31 & 24.89 & 37.86 \\
Qwen-7B-Proxy-Tuning & 39.83\dec{1.87} & 51.40\imp{1.88} & \textbf{43.30\imp{5.61}} & \textbf{46.97\imp{1.52}} & 29.69\imp{1.38} & 24.89\imp{0.00} & 40.79\imp{2.93}\\
\hline
\end{tabular}%
\caption{Proxy-Tuning for Larger Models}
\label{tab:proxy}
\end{table*}

\section{The Application of the Lite Apollo: Proxy-Tuning for Larger Models}

\paragraph{Preliminaries} Inspired by ~\cite{liu2024tuning,liu2021dexperts}, we introduce a lightweight model-agnostic decoding method in medical senarios. We leverage the logits from both pre and post fine-tuned small models to indirectly steer the larger base model's adjustments, thereby eschewing the need for direct parameter fine-tuning. Let $M_{raw}$ denote the smaller pre-trained model, and $M_{tuned}$ denote its fine-tuned counterpart. We compute the logit offset as "proxy" for each token, corresponding to the anti-expert and expert roles as delineated in ~\citet{liu2021dexperts}. This offset is then applied to the base model $M_{base}$ to synchronize the predictive distributions of the smaller and larger models. The modified probability distribution is given by:



{\small
$$
p'_{\mathcal{M}_{base}}(X_{t}\mid x_{1,...,t-1})=\text{softmax}\left[l_{M_{base}} + \Delta l_{M}\right]
$$

$$
\propto p_{\mathcal{M}_{base}}(X_{t}\mid x_{1,...,t-1}) \left(\frac{p_{\mathcal{M}_{tuned}}(X_{t}\mid x_{1,...,t-1})}{p_{\mathcal{M}_{raw}}(X_{t}\mid x_{1,...,t-1})}\right)
$$
}

where $\Delta l_{M} = l_{M_{tuned}} - l_{M_{raw}}$ represents the logit offset of the expert model $M_{tuned}$ and the anti-expert pre-trained model $M_{raw}$.The logit output for $M$ at $t$ is denoted by $l_{M_{t}}$ for the current timestep $t$. The probability distribution of $M$ refers to $p_{\mathcal{M}_{base}}(X_{t}\mid x_{1,...,t})$.

\paragraph{Settings}
$M_{base}$ is designated as the subject of investigation for Qwen-7B. Apollo-1.8B and Qwen-1.8B are appointed as the $M_{tuned}$ and $M_{raw}$.




\paragraph{Results}
As shown in the Tab. \ref{tab:proxy}, the overall effect of the model improves a lot without changing the parameters after proxy-tuning. From language perspective, except English, all other languages increase, and French has the most obvious increase. Excitingly, for French and Spanish, the model after proxy-tuning performs better than both $M_{tuned}$ and $M_{base}$, indicating that new accurate knowledge is generated after proxy-tuning. We also notice a decline in English proficiency. This may be because there is a gap between the distribution of difference and the probability itself, which leads to over-strengthening of the second option and requires further exploration and optimization.

\section{Related Work}


The integration of Large Language Models (LLMs) into the medical domain has sparked both enthusiasm and concern. These models demonstrate a remarkable ability to respond accurately to free-text queries using domain-specific knowledge. For instance, Google's Med-PaLM 2~\citep{singhal2023towards} stands out as the first medical LLM to achieve an expert level on the USMLE2-style questions in the MedQA dataset, boasting an accuracy exceeding 85$\%$. 

From a language perspective, many excellent works have appeared in the Chinese and English medical fields respectively. For Chinese, HuatuoGPT~\citep{chen2023huatuogpt} and BenTsao~\citep{wang2023huatuo} achieved good results by training on Chinese wikis, papers and medical consultation data. For English, Meditron~\citep{chen2023meditron} and PMC-LLaMA \citep{wu2023pmc} address limitations in medical knowledge accuracy of existing LLMs, tuning base models on millions of biomedical papers. For other languages, to the best of our knowledge, corresponding Medical LLMs have not yet emerged.





There have been some outstanding works focusing on multilingual topics recently. BioMistral~\citep{labrak2024biomistral} introduce the perspective of a multilingual evaluation system for the first time. MMedLM~\citep{qiu2024towards} is the first large medical model trained on multilingual corpus. We believe that our work, together with the formers, will bring a multilingual perspective into the medical artificial intelligence community and help more people with Medical AI.

\section{Conclusion}

In order to serve more people and larger community, we carefully collect and organize a high-quality medical corpus covering most populous languages in the world, open sourcing multi-language Dataset \textbf{ApolloCorpora} and evaluation set \textbf{XMedBench}. Based on these, we explore suitable methods for multilingual training and interrelationships between languages in the medical field, and finally obtains a series of models named  \textbf{Apollo}, with SOTA performance from 0.5B to 7B. Meanwhile,  proxy-tuning is  used to improve large foundation in terms of multilingual medical capabilities without changing the parameters.  We offer a \textbf{foundation} for global researchers, specially those with limited resources, to investigate medical LLMs.

\newpage
\section*{Acknowledgement}
This work was supported by  the Shenzhen Science and Technology Program (JCYJ20220818103001002), Shenzhen Doctoral Startup Funding (RCBS20221008093330065), Tianyuan Fund for Mathematics of National Natural Science Foundation of China (NSFC) (12326608), Shenzhen Key Laboratory of Cross-Modal Cognitive Computing (grant number ZDSYS20230626091302006), and Shenzhen Stability Science Program 2023, Shenzhen Key Lab of Multi-Modal Cognitive Computing.



\bibliography{custom}
\newpage
\appendix

\section{Details of ApolloCorpora, Multilingual Medical Dataset}
\label{sec:ApolloCorpora}

\subsection{Dataset Taxonomy and Collection of ApolloCorpora}

As shown in Tab. \ref{tab:data_tax}, we collect multilingual data from the data collection direction described in the first section of this chapter, which we will introduce in detail below.

\textbf{Books}
For English books, we use medical dictionary \footnote{\url{https://www.nlm.nih.gov/research/umls/new_users/online_learning/LEX_001.html}} to filter the books in the Pile Dataset~\citep{gao2020pile} and select books with medical words accounting for more than 4\%, and finally obtain 2312 medical-related books. For Chinese books, we follow MedQA~\citep{jin2020disease} to collect medical textbooks included in the five-year and eight-year medical student training programs in mainland China, and finally obtain 90 books.

 \textbf{Papers}
For English papers, we sample the public data in PubMed and obtain 878,241 medical abstracts. For Chinese papers, we also screen a total of 177,261 abstracts of papers published by the Chinese Medical Association \footnote{\url{https://www.yiigle.com/index}}. For French papers, we use the MORFITT~\citep{labrak:hal-04131591} dataset and the scientific article portion of the CLEAR~\citep{grabar2018clear}. For the Spanish paper, we use paper abstracts open sourced by the Mesinesp~\citep{gasco2021overview}.
\begin{figure*}[]\footnotesize
\begin{AIbox}{Prompts}

\textbf{Prompt for Generating Question: } \\
Please create a \textless question \textgreater that closely aligns with the provided \textless text\textgreater. Ensure that the \textless question\textgreater is formulated in English and does not explicitly reference the text. You may incorporate specific scenarios or contexts in the \textless question\textgreater, allowing the \textless text\textgreater to serve as a comprehensive and precise answer.
\\
\textless text\textgreater: \{text\}
\\
\textless question\textgreater: 
\\
\\
\textbf{Prompt for Generating Answer: } \\
You are Apollo, equipped with in-depth knowledge in medicine. Your task is to directly answer the user's \textless question\textgreater  in English. In formulating your response, you must thoughtfully reference the \textless reference text\textgreater, ensuring that your reply does not disclose your reliance on \textless reference text\textgreater. Aim to provide a comprehensive and informative response, incorporating relevant insights from \textless reference text\textgreater  to best assist the user. Please be cautious to avoid including any content that might raise ethical concerns.
\\
\textless question\textgreater: \{question\}
\\
\textless reference text\textgreater: \{reference\}
\textless reply\textgreater:

\end{AIbox} 
\caption{Prompts for Generating QA Pairs from Texts. We show the English version of Prompt, and other languages are similar.}
\label{fig:prompt_qa} 
\end{figure*}

\begin{figure*}[]\footnotesize
\begin{AIbox}{Prompt}
\textless text\textgreater  \{text\}\textless/text\textgreater \\
Please create some dialogues between patients and doctors in English based on the above text. The format is: \\
\textless Patient\textgreater Patient’s question\textless/Patient\textgreater \\
\textless Doctor\textgreater Doctor’s answer\textless/Doctor\textgreater \\
Both patient questions and doctor responses are as complex and detailed as possible.
\end{AIbox} 
\caption{Prompt Template for Generating Doctor-Patient Dialogues}
\label{fig:prompt_log} 
\end{figure*}

 \textbf{Encyclopedias}
For the English Encyclopedia, we also use the English Medical Dictionary to filter out 36107 medical-related wiki pages from dataset\footnote{\url{https://huggingface.co/datasets/wikipedia}}. For the French encyclopedia, we select the encyclopedia articles part of the CLEAR~\citep{grabar2018clear}. For the Hindi encyclopedia, we choose the HHD corpus~\citep{jain2018named}, which crawls descriptions of people, diseases, medical consumer products, and symptoms from Indian websites.

\textbf{Doctor-Patient Dialogues}
For Chinese, we directly use the HuatuoGPT dataset~\citep{zhang2023huatuogpt} and the simplified data set in Huatuo\_26M~\citep{li2023huatuo}. For English, we construct a multi-turn conversation data set based on PMC-Patients~\citep{zhao2022pmc} using ChatGPT, Prompt is shown in the Fig. \ref{fig:prompt_log}. For Arabic, we extract high-quality questions and answers with both question and answer lengths greater than 128 from the largest Arabic healthcare question and answer dataset MAQA~\citep{maqa}.

\textbf{Exams}
For the Chinese exam, we collect training sets of CMB~\citep{wang2023cmb}, CMExam~\citep{liu2024benchmarking}, and MedQA~\citep{zhang2018medical}. For the English exam, we collect the training sets of MedQA, Medmcqa~\citep{pal2022medmcqa} and Pubmedqa~\citep{jin2019pubmedqa}. For the Spanish and French exam, we select the training set of HEAD-QA~\citep{vilares2019head} and Frenchmcqa~\citep{labrak2023frenchmedmcqa} separately.

\textbf{Guidelines}
For English Guidelines, we select data from three sub-items of NICE\footnote{\url{https://www.nice.org.uk/guidance}}, PubMed and SPOR\footnote{\url{https://sporevidencealliance.ca/key-activities/cpg-asset-map/cpg-database/}} in the clinical guidelines introduced by Meditron~\citep{chen2023meditron}.


\textbf{General Instruction Tuning}
We use the translation~\citep{langsharegpt} and original data of Sharegpt\footnote{\url{https://github.com/lm-sys/FastChat}} and Alpaca~\citep{taori2023alpaca}. For Chinese, we additionally make use of data~\citep{wizardv2zh} generated by GPT-4 based on WizardLM Method~\citep{xu2023wizardlm}. For English, in addition to adding the WizardLM Dataset, we also add belebele~\citep{bandarkar2023belebele} to enhance multi-language reading comprehension capabilities, ai2\_arc~\citep{allenai:arc} to enhance abstract reasoning capabilities, Capybara~\citep{daniele2023amplify-instruct} to enhance instruction following capabilities.

 \textbf{Web}
For Chinese, we use the medical dictionary~\citep{meddiczh} to filter out medical-related articles from the Wudao Dataset~\citep{c6a3fe684227415a9db8e21bac4a15ab}. For English, we use the English Medical Vocabulary\footnote{\url{https://www.nlm.nih.gov/research/umls/new_users/online_learning/LEX_001.html}} to filter out medical related articles in C4 Dataset~\citep{2019t5}. For Spanish, we sampled 10\% of CoWeSe Dataset~\citep{carrino2021spanish}. \textbf{Math}
For mathematical abilities, we choose MathInstruct~\citep{yue2023mammoth}, a composite dataset containing various mathematics-related tasks and problem formats. \textbf{Code}
We choose Python-Alpaca~\citep{pythonalpaca} and Leetcode-ZH-11k~\citep{codezh} respectively to strengthen the ability to solve coding tasks in Chinese and English.

\begin{figure*}[]\footnotesize
\begin{AIbox}{Examples}
\textbf{Chinese}: 
Representatives of exogenous \textbf{wind-cold} and high fever include: (A) Yinqiao Powder, (B) Qingwen Baidu Decoction, (C) Dachengqi Decoctio. (D) Xiaobuhu Decoction \\

\textbf{Arabic}: 
\textit{\textbf{Patient}}: “\textbf{Peace be upon you}, I feel very intense friction in my body and head, redness in the skin or its color changing to pink, and I am almost cut off in my body. 
This all happens when I start making an effort, even if it is the smallest of things, even if my voice rises or I get angry. 
Thank you, and\textbf{ may God’s peace, mercy, and blessings be upon you}.” \\
\textit{\textbf{Doctor}}: “Often this is sensitivity due to \textbf{exertion and sweating}, and they are signs of a disease called \textbf{fever of the nile}, or a rash or allergy to sweating that appears in the \textbf{summer season} in particular. It is preferable to use a cold shower and apply the pimples with a \textbf{weakening ointment}. If the area is small, apply \textbf{Calcipotriene} ointment to it, take an anti-histamine medication such as \textbf{Polaramine}." \\

\textbf{Hindi}: 
Disease: \textbf{Epilepsy}. \textbf{Tulsi} reduces many diseases like a panacea. A large amount of anti-oxidants are found in Tulsi which cures free radicals in the brain. In case of any type of brain disease, if taken daily If 20 basil leaves are chewed and eaten, it is very effective. \textbf{Brahmi leaves}: Brahmi leaves grow around our homes especially where there is soil. It is round and curved in shape. It is taken daily. Consuming it chewed on an empty stomach not only strengthens the memory but also reduces epileptic seizures.\\

\textbf{Spanish}: 
Due to its \textbf{high mercury} content, the \textbf{Spanish Agency} for Food Safety and Nutrition recommends pregnant women not consume: A. Ripe soft cheeses, such as Brie or Camembert, B. Pâté or foie-gras, C. Calmette and Guérin bile vaccine., D. Raw sausages \\

\textbf{French}: 
What vaccinations are required in \textbf{France}? A.Measles-Mumps-Rubella, B.Human papillomavirus, C.Diphtheria-Tetanus-Poliomyelitis, D.Whooping cough, E.Calmette and Guérin bile vaccine.

\end{AIbox} 
\caption{Examples of local language characteristics in \textbf{ApolloCorpora}}
\label{fig:lang_data} 
\end{figure*}

\subsection{Details for Data Rewriting of ApolloCorpora}
\label{sec:rewrite}
We want to explore whether rewriting the original pre-training corpus into QA pairs in the context of continuing training can help increase its medical capabilities without destroying the original model's capabilities. We use ChatGPT\footnote{gpt-3.5-turbo-16k-0613} to generate questions and answers for a certain paragraph. For paragraph interception, we divide it according to the basic semantic units in the data set, such as sections in books and guides, paragraphs in website data, single wiki entry and abstracts of papers. For basic semantic units that are too long, we comprehensively consider the knowledge expression density of the language and subdivide different languages into blocks of different lengths to ensure that the semantic information covered by a single paragraph does not exceed the amount of information that can be included in a question and answer pair. For Spanish, French, English and Hindi we use 2048, for Chinese we use 256 and for Arabic we use 128. Prompts for generating QA pairs are detailed in the Fig. \ref{fig:prompt_log} and Fig. \ref{fig:prompt_qa}.



\subsection{Localized features of ApolloCorpora}
\label{sec:localization}

As shown in the Fig. \ref{fig:lang_data}, we illustrate the local language features in the dataset by language:

In terms of \textbf{symptom diagnosis}, local languages retain the terminology of traditional medicine, and due to different geographical environments and living habits, the possibility that a certain symptom corresponds to different diseases is also different: for Chinese, a disease has two aspects: "bìng" and "zhèng". The former is often translated as "disease entity". The latter, and more important one, is usually translated as "pattern". For example, the disease entity of a common cold might present with a pattern of wind-cold in one person, and with the pattern of wind-heat in another\footnote{\url{https://en.wikipedia.org/wiki/Traditional_Chinese_medicine\#Six_Excesses}}.

In terms of \textbf{medicines}, each language has its own specific names for medicines, and even retains some medicines from traditional medicine: for Chinese, there are about 13,000 medicines recorded in ancient Chinese literature and more than 100,000 Chinese medicine prescriptions; for Arabic and Hindi, doctors may also include some local plants in their medicines.

In terms of \textbf{communication terms}, some languages will have religious-related idioms at the beginning and end to improve the communication experience, such as Arabic.

In terms of \textbf{medical practice standards and dietary recommendations}, different medical systems have different standards, and different places also have different customary diets: for Spanish and French, local standards may differ, and dietary recommendations are also consistent with the preferences of the local population.


\section{Details of XMedBench}
\label{sec:bench}

\subsection{Construction of XMedBench}
\label{sec:bench_construct}
For English, we use the MedQA-USMLE~\citep{zhang2018medical}, MedMCQA~\citep{pal2022medmcqa}, and medical-related parts of MMLU~\citep{hendrycks2020measuring}; for Chinese, we used the of MedQA-MCMLE~\citep{zhang2018medical} and medical-related parts CMMLU~\citep{li2023cmmlu}; for Spanish, we used HEAD-QA~\citep{vilares2019head}; for French, we used FrenMedMCQA~\citep{labrak2023frenchmedmcqa}; For Arabic and Hindi, which lack local assessments, we make a compromise by applying translated versions of MMLU\footnote{Hindi: \url{https://huggingface.co/datasets/FreedomIntelligence/MMLU_Hindi}; Arabic: \url{https://huggingface.co/datasets/FreedomIntelligence/MMLU_Arabic}}. Specifically, we follow Med-PaLM2~\citep{singhal2023towards} and select six subcategories in MMLU: Clinical knowledge, Medical genetics, Anatomy, Professional medicine, College biology, and College medicine. For MedQA, we choose the 4-options version. For CMMLU, we select seven subdirectories: Anatomy, Clinical knowledge, College medicine, Genetics, Nutrition, Traditional chinese medicine, and Virology.



\subsection{Models for XMedBench}
\label{sec:bench_models}
\textbf{Qwen} Qwen is a suite of large language models from the Aliyun-developed Tongyi Qianwen from 0.5 billion to 72 billion parameters, based on the Transformer architecture and are trained on a diverse and extensive range of pretraining data. The types of pretraining data are varied and cover a wide scope, including a vast array of internet texts, professional books, code, and more.

 \textbf{Meditron}
Meditron is a suite of open-source medical large language models from 7 billion to 70 billion parameters, adapted to the medical domain from Llama-2 through continued pretraining on a comprehensively curated medical corpus, including selected PubMed articles, abstracts, a new dataset of internationally-recognized medical guidelines, and general domain data from RedPajama-v1.

 \textbf{Llama-2}
Llama-2 is a collection of pretrained and fine-tuned generative text models ranging in scale from 7 billion to 70 billion parameters. The fine-tuned versions use supervised fine-tuning (SFT) and reinforcement learning with human feedback (RLHF) to align to human preferences for helpfulness and safety.

 \textbf{Gemma}
Gemma is a family of lightweight, state-of-the-art open models from Google, built from the same research and technology used to create the Gemini models. They are text-to-text, decoder-only large language models, available in English, with open weights, pre-trained variants, and instruction-tuned variants. 

 \textbf{MMedLM2}
MMedLM 2 is a multilingual medical foundation model available in two versions, with parameter sizes of 1.8 billion and 7 billion. MMedLM 2 builds upon the foundation of InternLM 2 and has been further pretrained on MMedC, a comprehensive multilingual medical corpus. This further pretraining enhances the model's medical-domain knowledge.

 \textbf{Yi}
The Yi series models are the next generation of open-source large language models trained from scratch by 01.AI. Targeted as a bilingual language model and trained on 3T multilingual corpus, the Yi series models show promise in language understanding, commonsense reasoning, reading comprehension, and more. 

 \textbf{Mistral}
Mistral is a pretrained generative text model with 7 billion parameters. It uses Grouped-query attention (GQA) for faster inference and Sliding Window Attention (SWA) to handle longer sequences at smaller cost.

 \textbf{Zephyr}
Zephyr is a series of language models that are trained to act as helpful assistants, which is a fine-tuned version of mistralai/Mistral-7B-v0.1 that was trained on on a mix of publicly available, synthetic datasets using Direct Preference Optimization (DPO).

 \textbf{BioMistral}
BioMistral is a suite of Mistral-based further pre-trained open source models suited for the medical domains and pre-trained using textual data from PubMed Central Open Access. All the models are trained using the CNRS (French National Centre for Scientific Research) Jean Zay French HPC.

 \textbf{HuatuoGPT-2}
HuatuoGPT2 is a suite of open-source medical large language models from 7 billion to 34 billion parameters, which employs an innovative domain adaptation method to significantly boost its medical knowledge and dialogue proficiency. It showcases state-of-the-art performance in several medical benchmarks, especially surpassing GPT-4 in expert evaluations and the fresh medical licensing exams.

 \textbf{PMC-Llama}
MedLlama is initialized from Llama-13B and further pretrained with medical corpus. Despite the expert knowledge gained, it lacks instruction-following ability. It provides a instruction-tuning dataset and evaluates the tuned model.
MedLlama is pretrained on medical corpus, and PMC\_Llama is further finetuned based on MedLlama.

 \textbf{Apollo (Ours)}
Apollo is a suite of open-source medical large language models from 1.8 billion to 7 billion parameters.
The priority of all data items from the pre-training corpus to 16, and the priority of all data items from the instruction tuning stage to 2. The Batch size of model training is set to 256, the learning rate is set to 1e-4 for most models and 1e-5 for 7B model, and the warm up rate of Cosine scheduler is set to 0.03. The pre-training corpus is trained for one epoch, the Instrument Tuning corpus is trained for two epochs.

\section{Settings of Proxy-Tuning}
We set the priority of all data items from the pre-training corpus to 16, and the priority of all data items from the instruction tuning stage to 2. The Batch size of model training is set to 256, the learning rate is set to 1e-4, and the warm up rate of Cosine scheduler is set to 0.03. The pre-training corpus is trained for one epoch, the Instrument Tuning corpus is trained for two epochs.

\end{document}